\documentclass[
]{ceurart}

\usepackage{listings}
\usepackage{subcaption}
\usepackage[T1]{fontenc}
\usepackage{lmodern}

\definecolor{mgreen}{RGB}{114, 182, 161}
\definecolor{morange}{RGB}{233, 150, 117}
\definecolor{mblue}{RGB}{149, 163, 195}
\definecolor{mpink}{RGB}{219, 159, 192}

\begin{document}

%%
%% Rights management information.
%% CC-BY is default license.
\copyrightyear{2024}
\copyrightclause{Copyright for this paper by its authors.
  Use permitted under Creative Commons License Attribution 4.0
  International (CC BY 4.0).}

%%
%% This command is for the conference information
\conference{ }

%%
%% The "title" command
\title{Exploring Zero-Shot Anomaly Detection with CLIP in Medical Imaging: Are We There Yet?}

\author[1]{Aldo Marzullo}[%
orcid=0000-0002-9651-7156,
email=aldo.marzullo@humanitas.it
]
\cormark[1]
\address[1]{IRCCS Humanitas Research Hospital - via Manzoni 56, 20089 Rozzano, Milan, Italy}

\author[1]{Marta Bianca Maria Ranzini}[%
orcid=0000-0001-8275-6028,
email=marta.ranzini@humanitas.it
]

%% Footnotes
\cortext[1]{Corresponding author.}

%%
%% The abstract is a short summary of the work to be presented in the
%% article.
\begin{abstract}
  Zero-shot anomaly detection (ZSAD) offers potential for identifying anomalies in medical imaging without task-specific training. In this paper, we evaluate CLIP-based models, originally developed for industrial tasks, on brain tumor detection using the BraTS-MET dataset. Our analysis examines their ability to detect medical-specific anomalies with no or minimal supervision, addressing the challenges posed by limited data annotation. While these models show promise in transferring general knowledge to medical tasks, their performance falls short of the precision required for clinical use. Our findings highlight the need for further adaptation before CLIP-based models can be reliably applied to medical anomaly detection.
\end{abstract}

%%
%% Keywords. The author(s) should pick words that accurately describe
%% the work being presented. Separate the keywords with commas.

%%
%% This command processes the author and affiliation and title
%% information and builds the first part of the formatted document.
\maketitle

\section{Introduction}
Anomaly detection (AD) in medical imaging plays a critical role in identifying and diagnosing diseases, often detecting rare or subtle anomalies that may go unnoticed by human observers \cite{tschuchnig2022anomaly}. In particular, zero-shot anomaly detection (ZSAD), which aims to identify anomalies without specific training on abnormal samples, holds great promise for medical applications where obtaining labeled data is both challenging and costly. Despite the progress in anomaly detection, the majority of current models are trained on large, domain-specific datasets, which limits their applicability to novel tasks and categories—especially in the medical field, where training data encompassing real world variability are often difficult to obtain \cite{segato2020artificial}.
Recently, models based on Contrastive Language-Image Pretraining (CLIP) have shown remarkable success in zero- and few-shot tasks across various domains \cite{radford2021learning}. These models, trained on vast amounts of diverse, publicly available image-text pairs, excel at generalizing to unseen categories with minimal task-specific fine-tuning. However, their effectiveness in the medical domain, where anomaly detection often involves identifying subtle, domain-specific abnormalities, remains underexplored \cite{zhao2023clip}.

In this paper, we explore the potential of CLIP-based models for medical anomaly detection by focusing on a brain tumor detection task using the BraTS dataset \cite{moawad2023brain}. While CLIP-based models have demonstrated superior performance in industrial AD tasks, it is unclear whether these models are capable of achieving similar success in the medical domain, where higher accuracy and domain specificity are critical for clinical applications. We compare the performance of several CLIP-based models on the BraTS dataset, aiming to assess whether these zero-shot models are ready for medical anomaly detection or if further domain adaptation is required. Our findings reveal that, while CLIP models show promise in transferring knowledge from general tasks to medical imaging, their performance in detecting anomalies such as brain tumors is not yet sufficient for clinical use. Therefore, significant improvements and adaptations are needed to fully harness the potential of CLIP-based models in medical anomaly detection, especially for tasks that require high sensitivity and precision.

\section{Methods}
\label{sec:methods}

\subsection{Dataset Description}
% https://www.synapse.org/Synapse:syn51156910/wiki/622553
We tested selected CLIP-based abnormality detection models to detect and segment brain metastases. This task was chosen due to the clinical significance of brain metastases, which are common secondary tumors that pose significant challenges in diagnosis and treatment. They are indeed characterized by an extreme variability in terms of lesion size, shape and localisation within the brain, thus representing an optimal case study for AD. To this aim, we used the the BraTS 2023 Brain Metastases \cite{moawad2023brain}. It consists of a retrospective collection of treatment-naive brain metastases mpMRI scans obtained from various institutions. The dataset includes pre-contrast T1-weighted (t1w), post-contrast T1-weighted (t1c), T2-weighted (t2w), and T2-weighted FLAIR (t2f) sequences. All data underwent standardized preprocessing, including conversion to NIfTI format, co-registration, resampling to 1mm\(^3\) resolution, and skull-stripping. Imaging volumes were manually segmented and refined by neuroradiologists. 

\smallskip

For this work, we used only the BraTS 2023 training set (165 patients), splitting 70\% for training and 30\% for testing. The whole tumor ground truth mask (WC = Nonenhancing tumor core + Surrounding non-enhancing FLAIR hyperintensity + Enhancing Tumor) was used as the target mask. As a case study, only the axial view of t2w images was considered. CLIP's preprocessing standardization was applied using OpenAI's ImageNet mean and standard deviation on each slice separately. Data augmentation techniques were employed to reduce overfitting during training.

\subsection{Zero Shot Anomaly Detection using CLIP}
CLIP (Contrastive Language-Image Pretraining) is a large-scale vision-language model that has shown remarkable success in zero-shot image classification tasks \cite{radford2021learning}. CLIP learns joint embeddings of images and text by aligning them in a shared feature space. This makes it particularly powerful for tasks requiring minimal supervision, as it can match images to text prompts without the need for training on specific labeled datasets. While CLIP's ability to generalize across domains has been demonstrated in various vision tasks, its capacity for anomaly detection, particularly in medical contexts, remains an open question.

\smallskip

\noindent ZSAD aims to identify anomalous patterns in images from categories that are not present during training. Given an image $I \in \mathbb{R}^{H \times W \times 3}$, the objective is to compute an image-level anomaly score $S \in [0, 1]$ and a pixel-level anomaly map $M \in \mathbb{R}^{H \times W}$, where larger values indicate a higher likelihood of an anomaly. Unlike traditional anomaly detection methods, ZSAD does not rely on training data from the target categories but instead leverages a pre-trained vision-language (VL) model such as CLIP, which has been trained on natural image-text pairs. The VL model encodes both image and text features. To perform ZSAD, textual prompts describing normal and anomalous states are commonly used, e.g., ``A photo of a normal \{object\}'' or ``A photo of a damaged \{object\},'' where \{object\} is replaced with the category of interest. The model computes cosine similarities between the image embeddings $\mathbf{F}_{\text{img}}$ and text embeddings $\mathbf{F}_{\text{text}}$ for normal and anomalous states. The pixel-level anomaly map $M$ is defined by comparing the similarity between image patch embeddings $\mathbf{F}_{\text{patch}}$ and the text embeddings for both normal ($\mathbf{F}_{\text{normal}}$) and anomalous ($\mathbf{F}_{\text{anomalous}}$) states:
\[
M_{i,j} = \frac{\exp(\cos(\mathbf{F}_{\text{patch}}^{(i,j)}, \mathbf{F}_{\text{anomalous}}))}{\exp(\cos(\mathbf{F}_{\text{patch}}^{(i,j)}, \mathbf{F}_{\text{normal}})) + \exp(\cos(\mathbf{F}_{\text{patch}}^{(i,j)}, \mathbf{F}_{\text{anomalous}}))},
\]
where $\cos(\cdot, \cdot)$ denotes the cosine similarity, and $i,j$ refer to the spatial location of the patch. The image-level anomaly score $S$ is then computed by aggregating the anomaly map $M$ to provide an overall score for the image.

\medskip

Several variations of the above described approach have been developed, introducing features such as object-agnostic and learnable text prompts or value-wise attention mechanisms for fine-grained localization. In these cases, a set of auxiliary data $\mathcal{D}_{\text{train}} = \{(I_i, G_i)\}_{i=1}^{N}$, where $I_i$ represents the images and $G_i \in \{0, 1\}^{H \times W}$ are the corresponding ground-truth masks for anomalies, is typically available to train the adaptation layers. In this paper, we evaluate four AD methods that adapt CLIP for zero-shot and few-shot anomaly detection in medical imaging. 

\begin{itemize}
    \item  AnomalyCLIP \cite{zhou2023anomalyclip}: This method leverages CLIP's vision-language alignment for zero-shot anomaly detection (ZSAD) across diverse domains. AnomalyCLIP introduces learnable object-agnostic text prompts to capture generic notions of normality and abnormality, allowing it to focus on detecting anomalies regardless of foreground object semantics. By ignoring object class labels and concentrating on abnormal regions, AnomalyCLIP aims to improve ZSAD performance in highly variable datasets, such as defect inspection and medical imaging.
    
    \item VAND \cite{chen2023april}: Developed for the Visual Anomaly and Novelty Detection (VAND) challenge, this method augments CLIP with additional trainable linear layers to map image features into the joint embedding space, enabling better alignment with text features. For the zero-shot setting, VAND compares the features of test images with reference images stored in memory banks, which improves anomaly detection accuracy. This method showed excellent results in industrial settings, winning the zero-shot track of the VAND challenge, but its performance on medical tasks, such as brain tumor detection, remains to be seen.

    \item AnoVL \cite{deng2023anovl}: AnoVL addresses CLIP's limitation in capturing fine-grained, patch-level anomalies by introducing a value-wise attention mechanism within the visual encoder. This mechanism enhances the ability to localize anomalies at the pixel level. Additionally, AnoVL utilizes domain-aware state prompting to refine the matching between visual anomalies and abnormal state text prompts. Further improvements are made using a test-time adaptation (TTA) technique that refines the anomaly localization results by fine-tuning lightweight adapters based on pseudo-labels generated by AnoVL itself.

    \item AdaCLIP \cite{cao2024adaclip}: AdaCLIP enhances CLIP for zero-shot anomaly detection by introducing hybrid learnable prompts, combining static prompts shared across images with dynamic prompts generated for each test image. This hybrid approach improves adaptability and generalization across diverse anomaly categories. AdaCLIP also integrates a Hybrid Semantic Fusion (HSF) module, boosting both pixel- and image-level detection accuracy.
\end{itemize}

In addition to using the standard CLIP backbone, we also leverage a specialized version of CLIP for medical imaging, called PMC-CLIP \cite{lin2023pmc}. PMC-CLIP is pretrained on the PMC-OA dataset, which contains 1.6 million image-caption pairs from PubMedCentral's OpenAccess subset, covering a wide range of biomedical modalities and diseases. This biomedical-specific pretraining may potentially be more suitable for tasks such as brain metastasis detection than the original CLIP model.
Notably, both CLIP and PMC-CLIP are designed and trained to process 2D images. To the best of our knowledge and at the time of writing, there is no publicly available CLIP-like model for 3D anomaly detection. This might represent a limitation of such 2D approaches in MR anomaly detection, as they would not be able to exploit the rich three-dimensional information for a more accurate and spatially-coherent identification of anomalies. With our experiments we also aim to investigate the impact on volume-wise (3D) anomaly detection while using a 2D approach.

\subsection{Experiments}
In this section, we describe the benchmark designed to evaluate four CLIP-based anomaly detection (AD) methods—VAND, AnomalyCLIP, AnoVL and AdaCLIP — on the brain tumor segmentation task using the BraTS dataset. The benchmark is designed to assess the capability of these methods, originally developed for industrial AD tasks, in detecting medical anomalies such as brain tumors. We explore different training strategies to test the generalization of these models across domains, including industrial, medical, and brain tumor datasets. More in detail, we designed four experimental setups:

\begin{itemize}
    \item (Industrial) Pretrained on Industrial AD Dataset: In the first setup, we evaluate each method in its original form, using models pretrained on an industrial anomaly detection dataset. This setup allows us to examine the generalization capabilities of these models in detecting medical anomalies such as brain tumors without any additional medical-specific training.
    \item (Finetune) Finetuned on BraTS: Next, we finetune the models using the BraTS dataset, which contains labeled examples of brain tumors. This setup assesses the improvement in performance when the models are adapted to a specific medical domain through supervised learning. In this setting, the CLIP model weights are frozen. Only the adaptive layers introduced by each modification are finetuned on BraTS. 
    \item (Brats) Training from scratch: In this experiment, we train the model in its original form, with a randomly initialized adapter layer, on the BraTS dataset. As in the previous setting, the CLIP model weights are frozen.
    \item (PMC) Using PMC-CLIP \cite{lin2023pmc} as backbone and training from scratch on BraTS: Finally, we replace the CLIP model pretrained on industrial data with a CLIP model pretrained on the PMC dataset. As in the third setting, the adaptive layers on each anomaly-detection approach are trained from scratch on the BraTS dataset. This setup explores whether pretraining on a medical dataset followed by further adaptation on the target medical task enhances the models' ability to detect brain tumors.
\end{itemize}

These methods are designed for 2D image processing. Since brain MRIs are 3D volumes, we treat each slice (axial view) as a 2D image and reconstruct the 3D anomaly map by stacking the 2D anomaly maps. For all four setups, we measure the performance of each method using Dice score (equivalent to the F1-score) as well as sensitivity, specificity, and positive predictive value (PPV). These metrics are computed for each 3D volume and averaged across all test subjects. Additionally, we compute the Area Under the Receiver Operating Characteristic curve (AUROC) and the maximum F1-score (F1-max) \cite{jeong2023winclip}, which are commonly used metrics for assessing segmentation quality in AD tasks \cite{zhou2023anomalyclip,cao2024adaclip}. Note that to calculate pixel-wise metrics, the AD problem is framed as a segmentation task by thresholding the 3D anomaly maps to generate binary masks.
In all experiments, we used the open-source implementations provided by the authors of each method. We kept the default settings and hyperparameters unchanged for training (e.g., number of epochs and batch size), except for the image size, which was set to $240 \times 240$. For AdaCLIP specifically, due to the significant time required for training and inference, we randomly sampled 50 patients for training (same size as the test set). Notice that, due to its specific training-free design, AnoVL was only evaluated on a single setting (namely \textit{industrial}). More details for models training are reported in Appendix Table \ref{tab:model_settings}.

\section{Results}

\subsection{Overall Performance Comparison}
\begin{figure}[h]
    \centering
    \includegraphics[width=\textwidth]{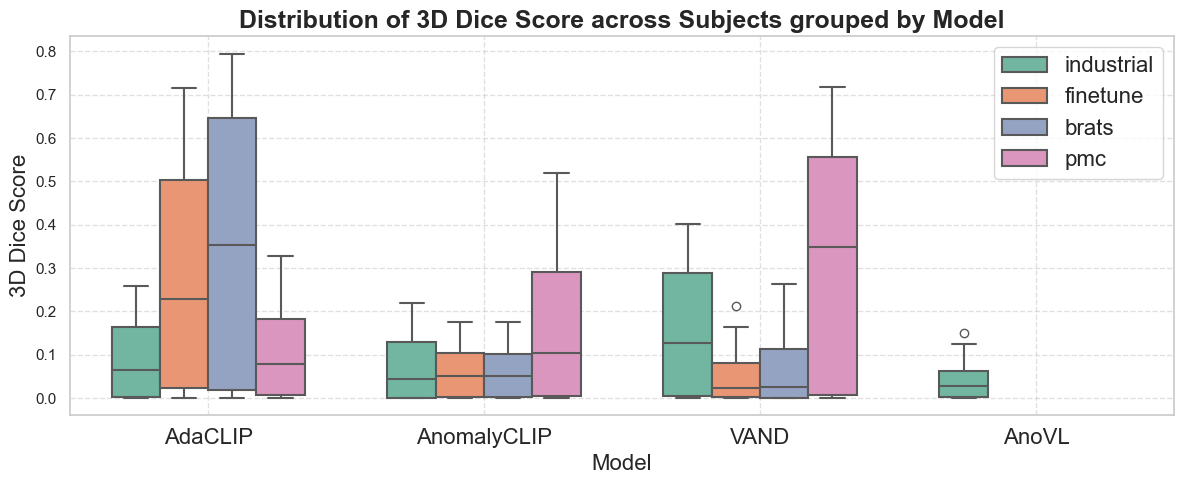}
    \caption{Distribution of 3D Dice scores across subjects grouped by AD method and for each training setup: \colorbox{mgreen}{\textbf{industrial}}, \colorbox{morange}{\textbf{finetune}}, \colorbox{mblue}{\textbf{brats}}, and \colorbox{mpink}{\textbf{pmc}}.}
    \label{fig:dice3d}
\end{figure}

The 3D Dice scores across subjects and models are summarized in Fig.~\ref{fig:dice3d}. This plot compares the performance of four anomaly detection models (AdaCLIP, AnomalyCLIP, VAND, and AnoVL) for each of the four experimental setups (pretrained on industrial datasets, finetuned on BraTS, training on BraTS from scratch, and training on BraTS using the PMC-CLIP backbone). 

\noindent AdaCLIP demonstrates the highest variance in performance across experiments, with its 3D Dice Scores ranging from approximately 0.1 to 0.7, particularly better when training the adapters on the BraTS dataset. VAND also shows relatively high performance, especially when using the PMC-CLIP backbone, where its scores vary between 0.1 and 0.6. In contrast, AnomalyCLIP shows consistently lower Dice Scores across all experiments, particularly struggling with scores close to 0 when using the \textit{industrial} and \textit{industrial-finetuned} weights. AnoVL shows the weakest performance overall, with very low scores across all experiments, rarely exceeding 0.1, reflecting its poor capacity for anomaly detection in this setting.

For further comparison, we analyze additional metrics as reported in Table \ref{tab:overall_results}. Notably, the trend observed in the AUROC values contrasts sharply with the Dice scores. We attribute this discrepancy to the AUROC's sensitivity to data imbalance \cite{davis2006auroc}, which can significantly affect the analysis of small lesions within large 3D volumes. Consequently, AUROC values may provide an inflated representation of performance that does not accurately reflect segmentation quality.

\begin{table}[htbp]
    \centering
    \caption{Overall performance of tested AD models on brain tumor segmentation task across different training setups. Metrics are averaged ($\pm$ standard deviation) over all test cases (3D images), except for AUROC and f1-max which are computed across all pixels of all test images, as in related literature.}
    \label{tab:overall_results}
    \resizebox{\textwidth}{!}{%
    \begin{tabular}{cc|cccccc}
        \toprule
        \textbf{Model} & \textbf{Training Setup} & \textbf{Dice Score} & \textbf{Sensitivity} & \textbf{Specificity} & \textbf{PPV} & \textbf{AUROC} & \textbf{F1-max} \\
        \midrule
        AdaCLIP & Industrial & 0.09 ± 0.08 & 0.16 ± 0.12 & 0.98 ± 0.00 & 0.07 ± 0.07  & 0.48 & 0.11\\
        AdaCLIP & Finetune & 0.29 ± 0.26 & 0.40 ± 0.24 & 0.99 ± 0.00 & 0.27 ± 0.26 & 0.66 & 0.46\\
        AdaCLIP & Brats & 0.35 ± 0.30 & 0.46 ± 0.31 & 0.99 ± 0.00 & 0.31 ± 0.28 & 0.35 & 0.58\\
        AdaCLIP & PMC & 0.11 ± 0.11 & 0.65 ± 0.24 & 0.91 ± 0.03 & 0.06 ± 0.07 & 0.90 & 0.15\\
        \hline
        AnoVL & Industrial & 0.04 ± 0.04 & 0.98 ± 0.03 & 0.63 ± 0.05 & 0.02 ± 0.02 & 0.91 & 0.11\\
        \hline
        AnomalyCLIP & Industrial & 0.07 ± 0.07 & 0.16 ± 0.13 & 0.97 ± 0.01 & 0.05 ± 0.06 & 0.63 & 0.08 \\
        AnomalyCLIP & Finetune & 0.06 ± 0.06 & 0.25 ± 0.18 & 0.96 ± 0.01 & 0.04 ± 0.04 & 0.42 & 0.06 \\
        AnomalyCLIP & Brats & 0.06 ± 0.06 & 0.25 ± 0.19 & 0.96 ± 0.01 & 0.04 ± 0.04 & 0.43 & 0.06\\
        AnomalyCLIP & PMC & 0.16 ± 0.17 & 0.35 ± 0.23 & 0.97 ± 0.01 & 0.12 ± 0.14 & 0.59 & 0.23\\
        \hline
        VAND & Industrial & 0.15 ± 0.14 & 0.51 ± 0.30 & 0.96 ± 0.01 & 0.09 ± 0.09 & 0.84 & 0.23\\
        VAND & Finetune & 0.05 ± 0.05 & 0.72 ± 0.20 & 0.73 ± 0.07 & 0.02 ± 0.03 & 0.83 & 0.41\\
        VAND & Brats & 0.06 ± 0.07 & 0.46 ± 0.33 & 0.83 ± 0.04 & 0.03 ± 0.04 & 0.44 & 0.06 \\
        VAND & PMC & 0.32 ± 0.26 & 0.55 ± 0.38 & 0.99 ± 0.00 & 0.24 ± 0.20  & 0.29 & 0.74\\
        \bottomrule
    \end{tabular}
    }
\end{table}

\subsection{Per-Subject Performance Analysis}
\begin{figure}[h]
    \centering
    \includegraphics[width=\textwidth]{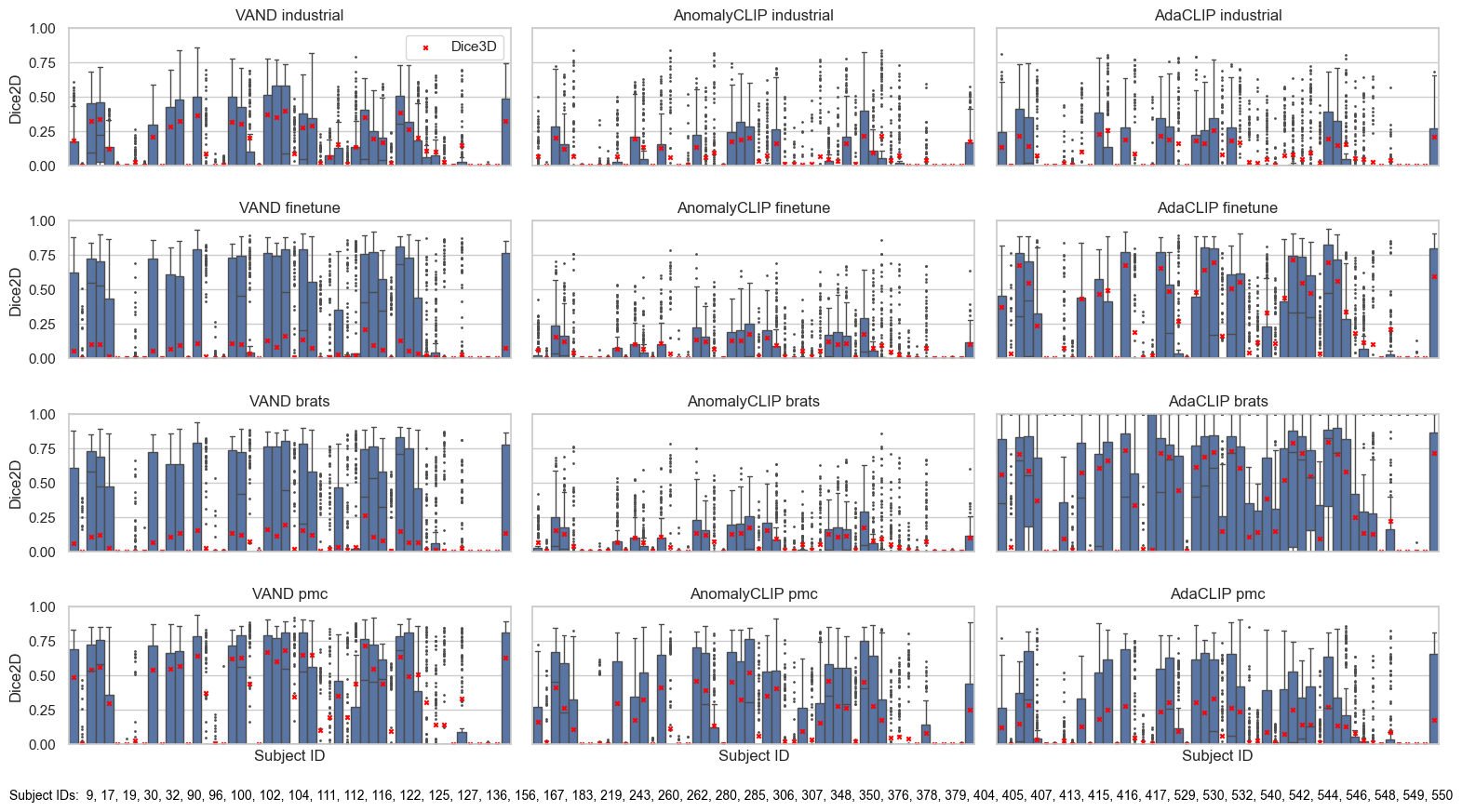}
    \caption{Boxplots of 2D Dice Scores for each subject across all models and datasets. Red crosses represent corresponding 3D Dice Scores for each subject. BraTS Subject IDs are listed at the bottom in the same order as they appear in the corresponding bar charts, from left to right.}
    \label{fig:dice2d_subject}
\end{figure}

To further investigate model performance, we analyzed Dice Scores per subject across the experiments, as shown in Fig.~\ref{fig:dice2d_subject}. Due to its low performance we exclude AnoVL from this analysis. Each plot presents 2D Dice Scores across individual slices (blue bars) and the corresponding 3D Dice Scores (red crosses). This comparison illustrates the extent of variability between 2D slice-level performance and 3D volumetric segmentation accuracy.

\begin{itemize}
    \item \textbf{VAND} exhibits relatively stable 2D segmentation accuracy across subjects, especially when trained on BraTS from scratch and when using PMC-CLIP backbone, where Dice Scores typically range from 0.4 to 0.8. However, the 3D Dice Scores are notably lower for most subjects, with values between 0.2 and 0.5, highlighting challenges in volumetric segmentation. To note, while showing similar variability of Dice score on 2D slices, the experiment on PMC-CLIP backbone reports higher 3D Dice scores compared to industrial-CLIP with training on BraTS, suggesting that, for this architecture, medical-specific CLIP embeddings are more suitable for the task of anomaly detection.  

    \item \textbf{AdaCLIP} also shows a wide range of 2D Dice Scores, particularly excelling in the training on BraTS, where it achieves values as high as 0.8 in certain subjects. However, like VAND, the corresponding 3D Dice Scores are consistently lower, ranging between 0.2 and 0.4 for many subjects.

    \item \textbf{AnomalyCLIP}, on the other hand, underperforms across all datasets. Similarly to VAND, the greatest improvement is obtained with the use of the PMC-CLIP and training the adaptive layers on BraTS. However, its 2D Dice Scores are generally below 0.3 for most subjects, and the 3D Dice Scores are even lower, barely exceeding 0.1 for the majority of cases. This suggests poor performance for anomaly detection in both 2D and 3D segmentations. 
\end{itemize}

\subsection{Error Distribution Across Brain Regions}
\begin{figure}[h]
    \centering
    \includegraphics[width=\textwidth]{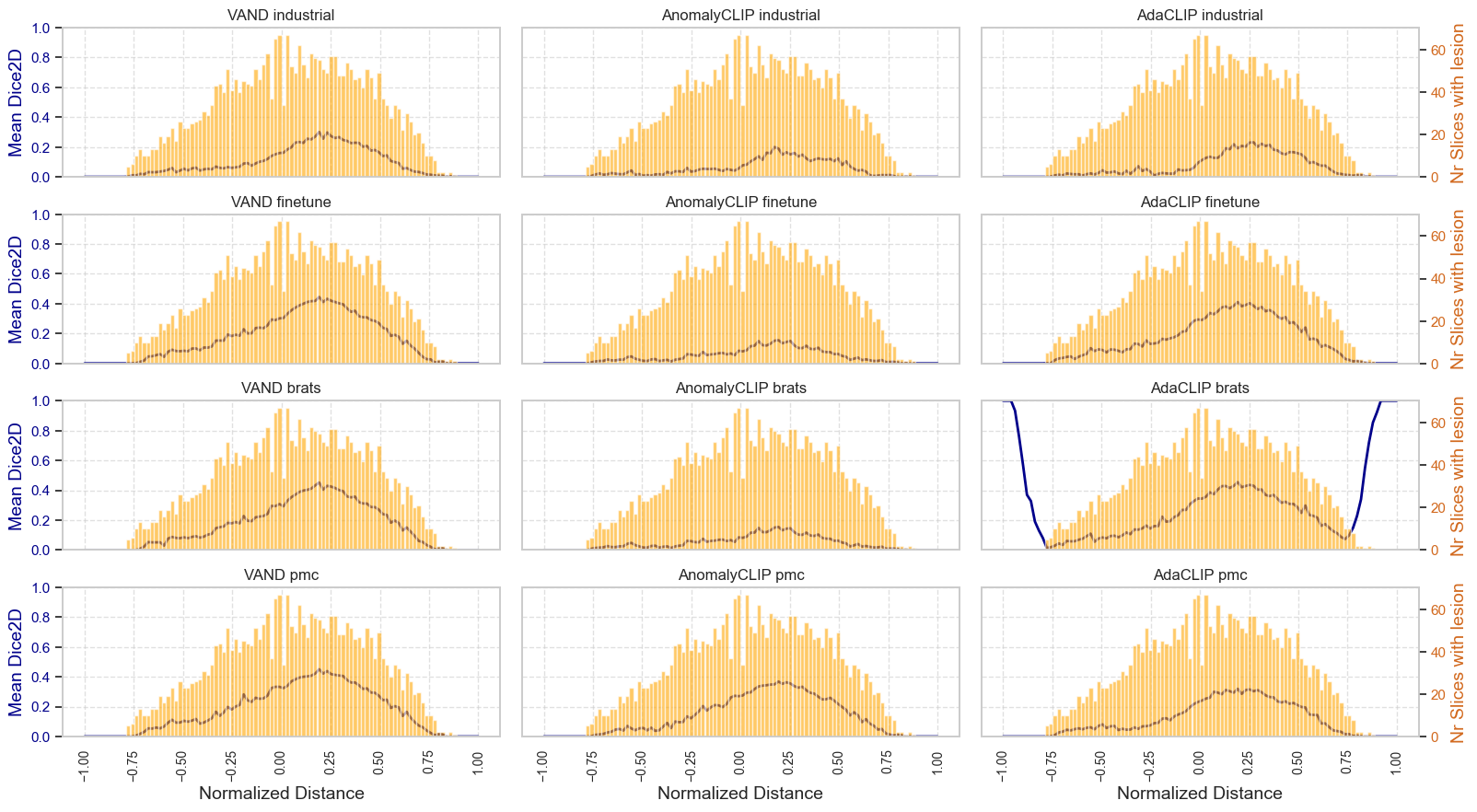}
    \caption{Mean 2D Dice Scores plotted against normalized distance from the center of the brain (0). A value of -1 corresponds to the inferior part of the brain, while 1 represents the superior part. Yellow bars indicate the number of slices with lesions at each distance.}
    \label{fig:dice2d_distance}
\end{figure}

To understand how segmentation performance changes spatially within the brain, we plotted the mean 2D Dice Score for each discretized distance from the brain center (normalized from -1 to 1) in Fig.~\ref{fig:dice2d_distance}. The number of slices containing lesions at each distance is also shown (yellow bars), providing insight into how lesion location affects segmentation accuracy.

\noindent For both \textbf{VAND} and \textbf{AdaCLIP}, segmentation accuracy is highest near the upper part of the brain (normalized distance around 0.25), where the number of lesion-containing slices is also highest. Dice Scores reach values between 0.4 and 0.7 at these upper-central regions, while they drop to around 0.2 towards the periphery (distances near -1 and 1). This suggests that these models struggle more with peripheral brain regions, where lesion frequency is lower or more variable. \textbf{AnomalyCLIP}, however, despite showing a similar trend as the other methods, achieves lower performance across all distances overall, with mean Dice Scores consistently below 0.3. This suggests that the model is unable to effectively capture lesion characteristics regardless of their location in the brain.

In summary, there is a trend in achieving reasonable 2D segmentation performance near the center of the brain but struggle towards the periphery. Their 3D performance remains lower than their 2D scores, suggesting challenges in volumetric segmentation.

\subsection{Discussion and Conclusion}
The results from our experiments show consistently low performance for all models on the BraTS brain tumor segmentation task, with Dice scores below 50\%, even when using PMC-CLIP, a visual-language model pretrained on a medical dataset. Although a comprehensive quantitative comparison is challenging due to variations in medical applications and datasets, we observe that the AUROC values obtained in our experiments (Table \ref{tab:overall_results}) are generally lower than those reported in the related studies. Indeed, despite pixel-level Dice scores not being reported for Brain MRI, AdaCLIP presents AUROC values ranging from 0.772 to 0.904, while AnomalyCLIP reports values between 0.789 and 0.897 for ZSAD across other (2D) medical imaging applications (e.g. skin, colon, thyroid). Concerning Brain anomaly detection, their reported image level AUROC values exceed 90\%. However, such high values do not align with pixel-level metrics observed in our experiments, also due to the fact that AUROC might not be sensitive enough to capture anomaly segmentation quality. Therefore, the Dice score should be preferred as a more reliable metric for performance evaluation in this context, particularly given the sparsity of anomalous voxels.

Nevertheless, the observed discrepancy may be attributable to the nature of the BraTS dataset, which presents unique challenges such as greater variability in tumor appearances and diffuse boundaries compared to the possibly simpler brain MRI datasets used in related work experiments, e.g. BrainMRI~\cite{kanade2015brain}. This latter dataset contains only 2D axial slices of 15 patients that have glial brain tumors mostly located in the upper part of the brain, which may be an easier task compared to the detection of metastasis in a 3D setting. This also aligns with the observation that, as shown in Figure \ref{fig:dice2d_distance}, all tested algorithms report higher performance in the central-upper part of the brain, suggesting that these areas might be better captured by the CLIP embeddings with respect to the rest of the brain. Therefore, our findings suggest that CLIP-based models may struggle with more complex and heterogeneous medical anomalies thus highlighting the need for more specialized fine-tuning and adaptation in these domains.
\smallskip

\noindent One notable limitation of our study is the slow training process associated with the AdaCLIP model, which stems from the use of batch size 1. Due to computational constraints, the cohort used for training was sampled down, potentially affecting the robustness and comparability of the results. The inefficiency caused by the small batch size severely hampered the speed of convergence and made the fine-tuning process impractically slow. Additionally, another critical factor that may have contributed to the low performance of the models is the choice of the initial prompt. The prompts used for zero-shot anomaly detection in brain imaging may not have been optimally suited for identifying brain anomalies, such as metastases, which have unique and complex characteristics. Future research may involve more specialized prompt templates to better align with the medical domain. Finally, while PMC-CLIP has demonstrated promising results when fine-tuned for downstream medical tasks, it has been observed that it may lack strong zero-shot anomaly detection capabilities. Other models, such as BiomedCLIP \cite{zhang2023biomedclip}, may offer more suitable alternatives for ZSAD tasks, although technical challenges in adapting these models to existing architectures still need to be addressed.

\smallskip

\noindent Finally, several other CLIP-based models have already been designed for medical imaging, including MedicalCLIP \cite{hua2024medicalclip}, MediCLIP \cite{zhang2024mediclip} and MVFA-AD \cite{huang2024adapting}. While these models show promise, their evaluation has so far been limited to 2D medical datasets, leaving uncertainty about their applicability to more complex 3D medical imaging tasks, such as brain metastasis detection. In a real-world context, the detection of brain metastases from 3D MRI images presents significant challenges due to the intricacies of tumor shape, size, and location, as well as the variability in imaging protocols. Thus, it remains unclear whether these models can generalize effectively to such challenging datasets, suggesting the need for future work on adapting these methods to handle 3D data and validating their performance on tasks involving more complex anatomical structures. Such evaluation is the goal of our future work.

%%
%% Define the bibliography file to be used
\bibliography{bib}

\begin{thebibliography}{17}
\expandafter\ifx\csname natexlab\endcsname\relax\def\natexlab#1{#1}\fi
\providecommand{\url}[1]{\texttt{#1}}
\providecommand{\href}[2]{#2}
\providecommand{\path}[1]{#1}
\providecommand{\DOIprefix}{doi:}
\providecommand{\ArXivprefix}{arXiv:}
\providecommand{\URLprefix}{URL: }
\providecommand{\Pubmedprefix}{pmid:}
\providecommand{\doi}[1]{\href{http://dx.doi.org/#1}{\path{#1}}}
\providecommand{\Pubmed}[1]{\href{pmid:#1}{\path{#1}}}
\providecommand{\bibinfo}[2]{#2}
\ifx\xfnm\relax \def\xfnm[#1]{\unskip,\space#1}\fi
%Type = Inproceedings
\bibitem[{Tschuchnig and Gadermayr(2022)}]{tschuchnig2022anomaly}
\bibinfo{author}{M.~E. Tschuchnig}, \bibinfo{author}{M.~Gadermayr},
\newblock \bibinfo{title}{Anomaly detection in medical imaging-a mini review},
\newblock in: \bibinfo{booktitle}{Data Science--Analytics and Applications:
  Proceedings of the 4th International Data Science Conference--iDSC2021},
  \bibinfo{organization}{Springer}, \bibinfo{year}{2022}, pp.
  \bibinfo{pages}{33--38}.
%Type = Article
\bibitem[{Segato et~al.(2020)Segato, Marzullo, Calimeri, and
  De~Momi}]{segato2020artificial}
\bibinfo{author}{A.~Segato}, \bibinfo{author}{A.~Marzullo},
  \bibinfo{author}{F.~Calimeri}, \bibinfo{author}{E.~De~Momi},
\newblock \bibinfo{title}{Artificial intelligence for brain diseases: A
  systematic review},
\newblock \bibinfo{journal}{APL bioengineering} \bibinfo{volume}{4}
  (\bibinfo{year}{2020}).
%Type = Inproceedings
\bibitem[{Radford et~al.(2021)Radford, Kim, Hallacy, Ramesh, Goh, Agarwal,
  Sastry, Askell, Mishkin, Clark et~al.}]{radford2021learning}
\bibinfo{author}{A.~Radford}, \bibinfo{author}{J.~W. Kim},
  \bibinfo{author}{C.~Hallacy}, \bibinfo{author}{A.~Ramesh},
  \bibinfo{author}{G.~Goh}, \bibinfo{author}{S.~Agarwal},
  \bibinfo{author}{G.~Sastry}, \bibinfo{author}{A.~Askell},
  \bibinfo{author}{P.~Mishkin}, \bibinfo{author}{J.~Clark}, et~al.,
\newblock \bibinfo{title}{Learning transferable visual models from natural
  language supervision},
\newblock in: \bibinfo{booktitle}{International conference on machine
  learning}, \bibinfo{organization}{PMLR}, \bibinfo{year}{2021}, pp.
  \bibinfo{pages}{8748--8763}.
%Type = Article
\bibitem[{Zhao et~al.(2023)Zhao, Liu, Wu, Li, Wang, Teng, Liu, Li, Cui, Wang
  et~al.}]{zhao2023clip}
\bibinfo{author}{Z.~Zhao}, \bibinfo{author}{Y.~Liu}, \bibinfo{author}{H.~Wu},
  \bibinfo{author}{Y.~Li}, \bibinfo{author}{S.~Wang},
  \bibinfo{author}{L.~Teng}, \bibinfo{author}{D.~Liu}, \bibinfo{author}{X.~Li},
  \bibinfo{author}{Z.~Cui}, \bibinfo{author}{Q.~Wang}, et~al.,
\newblock \bibinfo{title}{Clip in medical imaging: A comprehensive survey},
\newblock \bibinfo{journal}{arXiv preprint arXiv:2312.07353}
  (\bibinfo{year}{2023}).
%Type = Article
\bibitem[{Moawad et~al.(2023)Moawad, Janas, Baid, Ramakrishnan, Jekel,
  Krantchev, Moy, Saluja, Osenberg, Wilms et~al.}]{moawad2023brain}
\bibinfo{author}{A.~W. Moawad}, \bibinfo{author}{A.~Janas},
  \bibinfo{author}{U.~Baid}, \bibinfo{author}{D.~Ramakrishnan},
  \bibinfo{author}{L.~Jekel}, \bibinfo{author}{K.~Krantchev},
  \bibinfo{author}{H.~Moy}, \bibinfo{author}{R.~Saluja},
  \bibinfo{author}{K.~Osenberg}, \bibinfo{author}{K.~Wilms}, et~al.,
\newblock \bibinfo{title}{The brain tumor segmentation (brats-mets) challenge
  2023: Brain metastasis segmentation on pre-treatment mri},
\newblock \bibinfo{journal}{arXiv preprint arXiv:2306.00838}
  (\bibinfo{year}{2023}).
%Type = Article
\bibitem[{Zhou et~al.(2023)Zhou, Pang, Tian, He, and
  Chen}]{zhou2023anomalyclip}
\bibinfo{author}{Q.~Zhou}, \bibinfo{author}{G.~Pang},
  \bibinfo{author}{Y.~Tian}, \bibinfo{author}{S.~He},
  \bibinfo{author}{J.~Chen},
\newblock \bibinfo{title}{Anomalyclip: Object-agnostic prompt learning for
  zero-shot anomaly detection},
\newblock \bibinfo{journal}{arXiv preprint arXiv:2310.18961}
  (\bibinfo{year}{2023}).
%Type = Article
\bibitem[{Chen et~al.(2023)Chen, Han, and Zhang}]{chen2023april}
\bibinfo{author}{X.~Chen}, \bibinfo{author}{Y.~Han},
  \bibinfo{author}{J.~Zhang},
\newblock \bibinfo{title}{April-gan: A zero-/few-shot anomaly classification
  and segmentation method for cvpr 2023 vand workshop challenge tracks 1\&2:
  1st place on zero-shot ad and 4th place on few-shot ad},
\newblock \bibinfo{journal}{arXiv preprint arXiv:2305.17382}
  (\bibinfo{year}{2023}).
%Type = Article
\bibitem[{Deng et~al.(2023)Deng, Zhang, Bao, and Li}]{deng2023anovl}
\bibinfo{author}{H.~Deng}, \bibinfo{author}{Z.~Zhang},
  \bibinfo{author}{J.~Bao}, \bibinfo{author}{X.~Li},
\newblock \bibinfo{title}{Anovl: Adapting vision-language models for unified
  zero-shot anomaly localization},
\newblock \bibinfo{journal}{arXiv preprint arXiv:2308.15939}
  (\bibinfo{year}{2023}).
%Type = Article
\bibitem[{Cao et~al.(2024)Cao, Zhang, Frittoli, Cheng, Shen, and
  Boracchi}]{cao2024adaclip}
\bibinfo{author}{Y.~Cao}, \bibinfo{author}{J.~Zhang},
  \bibinfo{author}{L.~Frittoli}, \bibinfo{author}{Y.~Cheng},
  \bibinfo{author}{W.~Shen}, \bibinfo{author}{G.~Boracchi},
\newblock \bibinfo{title}{Adaclip: Adapting clip with hybrid learnable prompts
  for zero-shot anomaly detection},
\newblock \bibinfo{journal}{arXiv preprint arXiv:2407.15795}
  (\bibinfo{year}{2024}).
%Type = Inproceedings
\bibitem[{Lin et~al.(2023)Lin, Zhao, Zhang, Wu, Zhang, Wang, and
  Xie}]{lin2023pmc}
\bibinfo{author}{W.~Lin}, \bibinfo{author}{Z.~Zhao},
  \bibinfo{author}{X.~Zhang}, \bibinfo{author}{C.~Wu},
  \bibinfo{author}{Y.~Zhang}, \bibinfo{author}{Y.~Wang},
  \bibinfo{author}{W.~Xie},
\newblock \bibinfo{title}{Pmc-clip: Contrastive language-image pre-training
  using biomedical documents},
\newblock in: \bibinfo{booktitle}{International Conference on Medical Image
  Computing and Computer-Assisted Intervention},
  \bibinfo{organization}{Springer}, \bibinfo{year}{2023}, pp.
  \bibinfo{pages}{525--536}.
%Type = Article
\bibitem[{Jeong et~al.(2023)Jeong, Zou, Kim, Zhang, Ravichandran, and
  Dabeer}]{jeong2023winclip}
\bibinfo{author}{J.~Jeong}, \bibinfo{author}{Y.~Zou}, \bibinfo{author}{T.~Kim},
  \bibinfo{author}{D.~Zhang}, \bibinfo{author}{A.~Ravichandran},
  \bibinfo{author}{O.~Dabeer},
\newblock \bibinfo{title}{Winclip: Zero-/few-shot anomaly classification and
  segmentation},
\newblock \bibinfo{journal}{arXiv preprint arXiv:2303.14814}
  (\bibinfo{year}{2023}).
%Type = Inproceedings
\bibitem[{Davis and Goadrich(2006)}]{davis2006auroc}
\bibinfo{author}{J.~Davis}, \bibinfo{author}{M.~Goadrich},
\newblock \bibinfo{title}{The relationship between precision-recall and roc
  curves},
\newblock in: \bibinfo{booktitle}{Proceedings of the 23rd international
  conference on Machine learning}, \bibinfo{year}{2006}, pp.
  \bibinfo{pages}{233--240}.
%Type = Article
\bibitem[{Kanade and Gumaste(2015)}]{kanade2015brain}
\bibinfo{author}{P.~B. Kanade}, \bibinfo{author}{P.~Gumaste},
\newblock \bibinfo{title}{Brain tumor detection using mri images},
\newblock \bibinfo{journal}{Brain} \bibinfo{volume}{3} (\bibinfo{year}{2015})
  \bibinfo{pages}{146--150}.
%Type = Article
\bibitem[{Zhang et~al.(2023)Zhang, Xu, Usuyama, Xu, Bagga, Tinn, Preston, Rao,
  Wei, Valluri et~al.}]{zhang2023biomedclip}
\bibinfo{author}{S.~Zhang}, \bibinfo{author}{Y.~Xu},
  \bibinfo{author}{N.~Usuyama}, \bibinfo{author}{H.~Xu},
  \bibinfo{author}{J.~Bagga}, \bibinfo{author}{R.~Tinn},
  \bibinfo{author}{S.~Preston}, \bibinfo{author}{R.~Rao},
  \bibinfo{author}{M.~Wei}, \bibinfo{author}{N.~Valluri}, et~al.,
\newblock \bibinfo{title}{Biomedclip: a multimodal biomedical foundation model
  pretrained from fifteen million scientific image-text pairs},
\newblock \bibinfo{journal}{arXiv preprint arXiv:2303.00915}
  (\bibinfo{year}{2023}).
%Type = Article
\bibitem[{Hua et~al.(2024)Hua, Luo, Qi, and Long}]{hua2024medicalclip}
\bibinfo{author}{L.~Hua}, \bibinfo{author}{Y.~Luo}, \bibinfo{author}{Q.~Qi},
  \bibinfo{author}{J.~Long},
\newblock \bibinfo{title}{Medicalclip: Anomaly-detection domain generalization
  with asymmetric constraints},
\newblock \bibinfo{journal}{Biomolecules} \bibinfo{volume}{14}
  (\bibinfo{year}{2024}) \bibinfo{pages}{590}.
%Type = Article
\bibitem[{Zhang et~al.(2024)Zhang, Xu, Qiu, Yan, Lang, and
  Zhou}]{zhang2024mediclip}
\bibinfo{author}{X.~Zhang}, \bibinfo{author}{M.~Xu}, \bibinfo{author}{D.~Qiu},
  \bibinfo{author}{R.~Yan}, \bibinfo{author}{N.~Lang},
  \bibinfo{author}{X.~Zhou},
\newblock \bibinfo{title}{Mediclip: Adapting clip for few-shot medical image
  anomaly detection},
\newblock \bibinfo{journal}{arXiv preprint arXiv:2405.11315}
  (\bibinfo{year}{2024}).
%Type = Inproceedings
\bibitem[{Huang et~al.(2024)Huang, Jiang, Feng, Zhang, Wang, and
  Wang}]{huang2024adapting}
\bibinfo{author}{C.~Huang}, \bibinfo{author}{A.~Jiang},
  \bibinfo{author}{J.~Feng}, \bibinfo{author}{Y.~Zhang},
  \bibinfo{author}{X.~Wang}, \bibinfo{author}{Y.~Wang},
\newblock \bibinfo{title}{Adapting visual-language models for generalizable
  anomaly detection in medical images},
\newblock in: \bibinfo{booktitle}{Proceedings of the IEEE/CVF Conference on
  Computer Vision and Pattern Recognition}, \bibinfo{year}{2024}, pp.
  \bibinfo{pages}{11375--11385}.

\end{thebibliography}

%%
%% If your work has an appendix, this is the place to put it.
\appendix
\section{experiment settings}
Table \ref{tab:model_settings} reports the main settings used for training AdaCLIP, AnoVL, AnomalyCLIP, and VAND models.

\begin{table}[h]
\centering
\caption{Main settings used for training AdaCLIP, AnoVL (TTA), AnomalyCLIP, and VAND models.}
\label{tab:model_settings}
\begin{tabular}{lcccc}
\toprule
\textbf{Setting}             & \textbf{AdaCLIP}            & \textbf{AnoVL}                & \textbf{AnomalyCLIP}          & \textbf{VAND}                 \\
\midrule
\textbf{Model}               & ViT-L-14-336                & ViT-B-16-plus-240             & ViT-L-14-336                   & ViT-L-14-336                  \\ \hline
\textbf{Batch size}          & 1                           & 1                             & 8                             & 8                             \\ \hline
\textbf{Image size}          & 240                         & 240                           & 240                           & 240                           \\ \hline
\textbf{Epochs}              & 5                         & 5                             & 15                            & 10                            \\ \hline
\textbf{Learning rate}       & 0.01                         & 0.001                           & 0.001                         & 0.001                         \\ \hline
\textbf{Feature layers}      & {[}3, 6, 9{]}                         & {[}3, 6, 9, 12{]}             & {[}6, 12, 18, 24{]}           & {[}6, 12, 18, 24{]}           \\ \hline
\textbf{Prompting depth}     & 4                           &                            &                            &                            \\ \hline
\textbf{Prompting type}      & Static + Dynamic                          &                            &                            &                            \\ \hline
\textbf{K-clusters}          & 20                          &                            &                            &                            \\ \hline
\bottomrule
\end{tabular}
\end{table}

\section{Variability of lesions at normalized distances.}
\begin{figure}
\centering
\begin{subfigure}[b]{\textwidth}
   \includegraphics[width=1\linewidth]{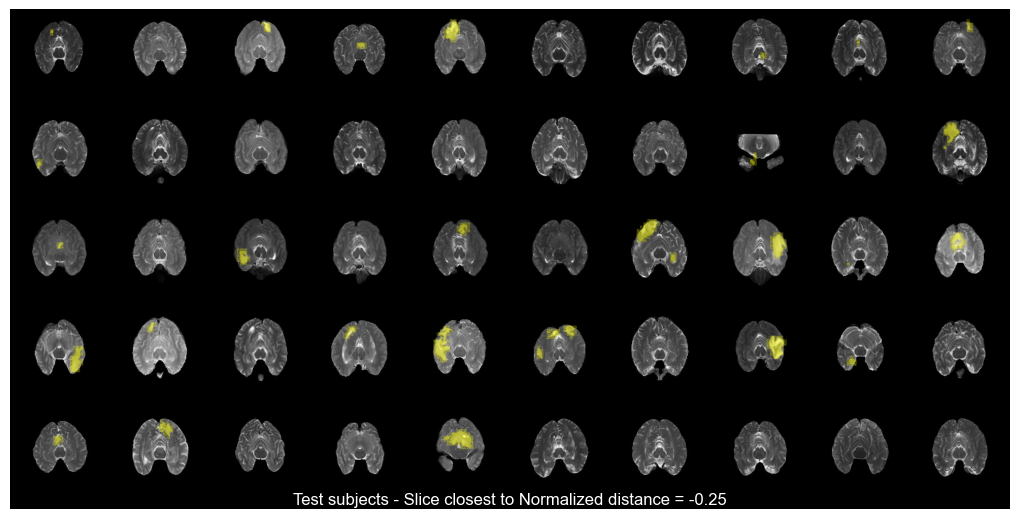}
   \caption{}
   \label{fig:lowersample} 
\end{subfigure}

\begin{subfigure}[b]{\textwidth}
   \includegraphics[width=1\linewidth]{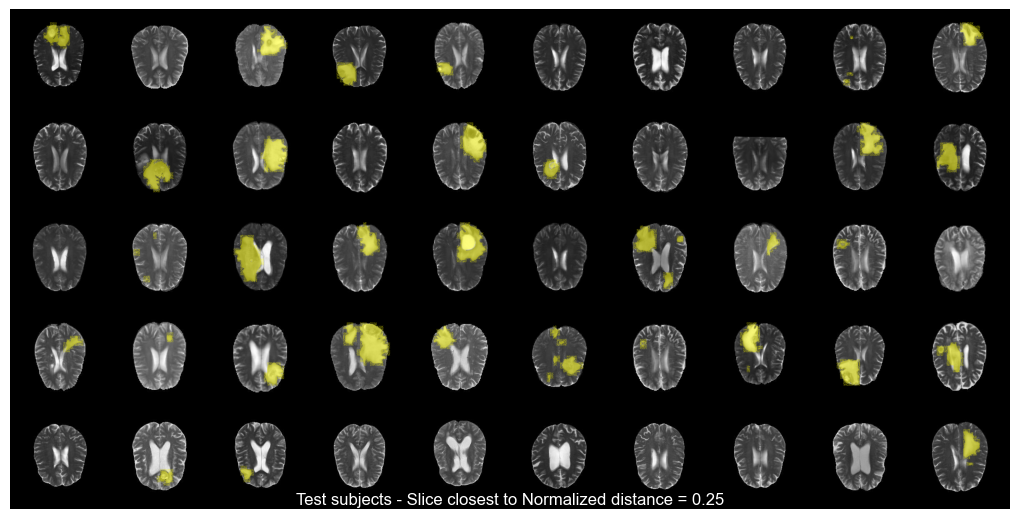}
   \caption{}
   \label{fig:uppersample}
\end{subfigure}

\caption{Sample 2D slices from 3D MRI images and their corresponding whole tumor segmentation (yellow overlap) for inferior part of the brain (a, distance $<$ 0) and superior part of the brain (b, distance $>$ 0).}
\end{figure}

\begin{figure}[h]
    \centering
    \includegraphics[width=\textwidth]{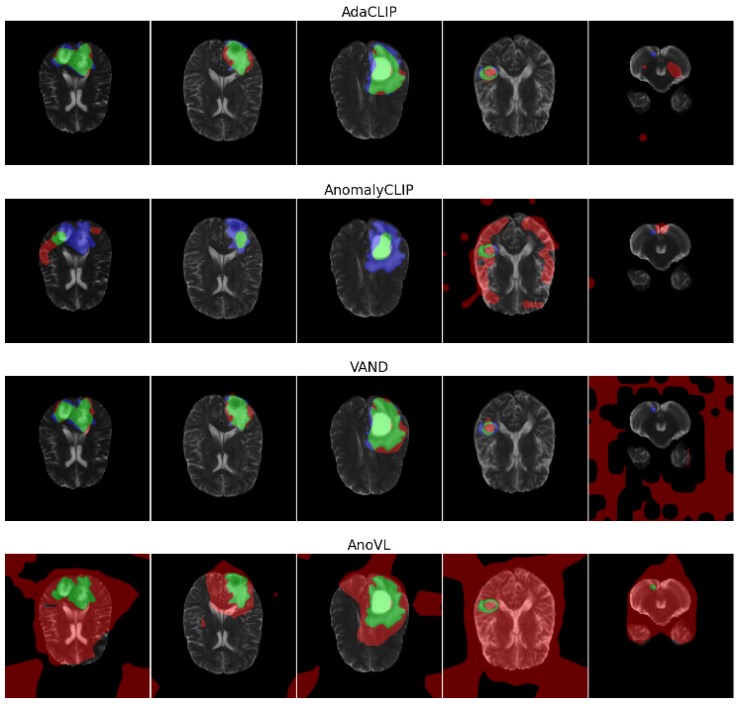}
    \caption{Overlay of ground truth and model predictions for MRI segmentation, highlighting true positives \colorbox{green}{(green)}, false positives \colorbox{red}{(red)}, and false negatives \colorbox{mblue}{(blue)}. Grayscale MRI slices provide anatomical context, showcasing model accuracy and errors. All the sample results refer to models finetuned on BraTS; better viewed in color.}
    \label{fig:samples_results}
\end{figure}

Figure \ref{fig:lowersample} and \ref{fig:uppersample} illustrate samples of 2D slices and their corresponding whole tumor segmentation for inferior and superior part of the brain, respectively. Notably, there is no significant correlation between lesion size and slice location, as indicated by a Spearman rank correlation coefficient ($r=0.08$). In contrast, moderate correlations can be observed between the size of the lesions and the Dice scores, as reported in Table \ref{tab:corrsize}.\\
\noindent Finally, Figure \ref{fig:samples_results} shows an overlay of ground truth and model predictions for MRI slice segmentation from randomly chosen subjects. This visualization, best viewed in color, illustrates the model's performance in identifying regions of interest and highlights areas of missed or over-segmented regions.

\begin{table}[h]
\centering
\caption{Spearman rank correlation coefficients between lesion size and Dice score.}
\label{tab:corrsize}
\begin{tabular}{lcccc}
\toprule
\textbf{Setting}             & \textbf{AdaCLIP}            & \textbf{AnoVL}                & \textbf{AnomalyCLIP}          & \textbf{VAND}\\
\midrule
\textbf{Industrial}    &   0.37    & 0.54    & 0.30    & 0.54                 \\ \hline
\textbf{Finetune}      &   0.46    & NA      & 0.36    & 0.54 \\ \hline
\textbf{Brats}         &   0.48    & NA      & 0.36    & 0.53 \\ \hline
\textbf{PMC}           &   0.51    & NA      & 0.44    & 0.55    \\ \hline
\bottomrule
\end{tabular}
\end{table}

\end{document}